\pdfoutput=1

\documentclass[11pt]{article}

\usepackage[]{acl}

\usepackage{times}
\usepackage{latexsym}
\usepackage[noalg]{stroop}
\usepackage{amsmath}
\usepackage[T1]{fontenc}

\usepackage[utf8]{inputenc}

\usepackage{microtype}

\usepackage{inconsolata}

\newcommand{\langpair}[2]{\texttt{#1-#2}}
\newcommand{\cmark}{\ding{51}}%
\newcommand{\xmark}{\ding{55}}%
\newcommand{\tbig}{\texttt{Transformer-big}}%
\newcommand{\tbase}{\texttt{Transformer-base}}%

\title{Representation Collapse in Machine Translation Through the Lens of Angular Dispersion.}

\author{\bf Evgeniia Tokarchuk, Maya K. Nachesa, Sergey Troshin \and Vlad Niculae \\
        Language Technology Lab, University of Amsterdam\\
        \texttt{evgeniia@tokarch.uk}, \texttt{\{m.k.nachesa,s.troshin,v.niculae\}@uva.nl}
}

\begin{document}
\maketitle
\begin{abstract}
Modern neural translation models based on the Transformer architecture are known for their high performance, particularly when trained on high-resource datasets. A standard next-token prediction training strategy, while widely adopted in practice, may lead to overlooked artifacts such as representation collapse. Previous works have shown that this problem is especially pronounced in the representation of the deeper Transformer layers, where it often fails to efficiently utilize the geometric space. Representation collapse is even more evident in end-to-end training of continuous-output neural machine translation, where the trivial solution would be to set all vectors to the same value. 
In this work, we analyze the dynamics of representation collapse at different levels of discrete and continuous NMT transformers throughout training. We incorporate an existing regularization method based on angular dispersion and demonstrate empirically that it not only mitigates collapse but also improves translation quality. Furthermore, we show that quantized models exhibit similar collapse behavior and that the benefits of regularization are preserved even after quantization.
\end{abstract}

\section{Introduction}
Text representations learned by Transformer-based models not only influence the performance of the target task, such as neural machine translation (NMT), but can also be extracted and used as a backbone for building retrieval-based models, such as retrieval-augmented generation \citep{Lewis2020RetrievalAugmentedGF}, $k$-Nearest Neighbors machine translation \citep{khandelwal2021nearest}, and continuous-output machine translation \citep{kumar2018von,tokarchuk-niculae-2022-target}. %
Effective training of Transformer models for sequential language tasks (such as NMT) is, however, known to be difficult,
in large part due to various forms of collapse of the internal representations learned
\citep{gao2018representation,godey-etal-2024-anisotropy,gerasimov2025fullyutilizetransformersrepresentation,barbero2024transformers,voita-etal-2019-bottom}.

\textit{Representation collapse} can be split into two phenomena: complete collapse and dimensional collapse \citep{hua-2021,jing2022understanding}. 
A complete collapse occurs when a trivial solution to the optimization problem is found by collapsing all vectors into a single point. It can be observed, for example, in continuous-output neural machine translation (CoNMT, introduced by \citet{kumar2018von} and further described in \cref{sec:nmt-collapse}), as well as in latent variable models \citep{pmlr-v119-chen20j}.
In contrast, dimensional collapse is partial, and occurs when a high-dimensional representation
space is underutilized and all representations end up lying in a lower-dimensional space.
While complete collapse is clearly visible from the training loss and automatic quality metrics, such as BLEU or COMET, dimensional collapse is not that evident,
but studies have captured it for decoder-only Transformers
\citep{barbero2024transformers}.

A popular approach to mitigate representation collapse in domains with continuous representation, such as images or videos, is to employ contrastive learning \citep{pmlr-v119-chen20j, jing2022understanding, pmlr-v119-wang20k}, which commonly relies on data augmentation and negative sampling to achieve diversity in the representations. Although data-augmentation of text data was also successfully applied in contrastive-learning setups \citep{wei-zou-2019-eda,Shen2020ASB,Su2021WhiteningSR,arefin2025seqvcr}, defining contrastive pairs for textual data is non-trivial due to the discrete nature of text. Despite being a competitive approach, one of the downsides of contrastive learning is that it can lead to higher variance of the objective function, in addition to typically requiring larger batch sizes to achieve a good performance \citep{guo2025representationlearningnoncontrastivemutual}. Moreover, performance depends heavily on the quality of the negative samples. \citet{pmlr-v119-wang20k} show that if we take into account the spherical geometry of the representation, contrastive learning, in fact, can be decomposed into two separate components. The first is \textit{alignment}, which enforces semantically close representations to be close in a vector space, and the second is \textit{uniformity}, which ensures better coverage of the space. Following this idea, we can focus on uniformity or angular dispersion \citep{tokarchuk2025keep} as a means to avoid representation collapse.

In this work, we focus on machine translation and revisit representation collapse in Transformer models. We analyze how collapse occurs in training and make a connection between representation collapse, the dimensionality of the model, and angular dispersion. We show that incorporating angular dispersion into the training of the model not only improves diversity of the representation but also leads to overall better MT quality.

\section{Background}
\subsection{Next Token Prediction for Machine Translation}\label{sec:nmt-collapse}
In NMT, given a source sequence \(\bm{x} = (x_1, \ldots, x_m)\), target sequence \(\bm{y}=(y_1, \ldots, y_n)\), and a finite vocabulary of type (sub)words $V=\{t_1,...,t_{|V|}\}$ such that $x_i,y_i \in V$, we aim to learn the mapping between the source and target sentences via the multi-class classification loss over $V$ which amounts to next-token prediction:
\begin{equation}\label{eq:discrete}
\begin{gathered}
L_\text{NMT}(y_i=t; \bm{y}_{<i}, \bm{x}) \\ = -\log p(y_i=t \mid \bm{y}_{<i}, \bm{x}).
\end{gathered}
\end{equation}
If we denote the latent representation produced by decoder as $\bm{H} = \bm{h}(\bm{x},\bm{y}_{<i})$ and embeddings of the vocabulary token $t$ as $\bm{E} = \bm{e}(t)$
\Cref{eq:discrete} can be expressed via Euclidean dot products as follows: 
\begin{equation}\label{eq:discrete-dot}
\begin{gathered}
-\log p(y_i=t \mid \bm{y}_{<i}, \bm{x}) \\ = -\DP{\bm{E}}{\bm{H}} + \log\sum_{t' \in V}\exp \DP{\bm{E'}}{\bm{H}},
\end{gathered}
\end{equation}
where \(t\) is a token index, \(V\) is the vocabulary, \(\bm{e} : V \to \bbR^d \)
is an embedding lookup,
and \(\bm{H}\) is a transformer hidden state calculated in terms of  \(\bm{x}\) and the output prefix \(\bm{y}_{<i}\).

The continuous alternative to next-token prediction, introduced by \citet{kumar2018von},
replaces classification with regression and
can be expressed using cosine similarity, which replaces the more costly log-sum-exp:
\begin{equation}\label{eq:cosineloss}
    L_\text{CoNMT}(y_i=t ; \bm{y}_{<i}, \bm{x})  = 1 - \operatorname{cos}(\bm{E}, \bm{H}).
\end{equation}
CoNMT was proposed as an efficient way to eliminate the training cost
of the last classification layer and log-sum-exp, which scales linearly with vocabulary size.
Instead, the regression loss does not depend on vocabulary size, but is vulnerable to collapse. 
All previous studies in this area use fixed embeddings, differing only in their specific configurations
\citep{kumar2018von, tokarchuk-niculae-2022-target, tokarchuk-niculae-2024-unreasonable}.

\subsection{Measure of Representation Collapse}
Previous works show that Transformer-based \citep{Vaswani-trafo} models trained with next token prediction loss are prone to dimensional representation collapse, which manifests as a lack of diversity in the tokens' representations \citep{arefin2025seqvcr, barbero2024transformers}. In the continuous case the matter gets even worse with complete collapse, \ie 
trivial global optima of \cref{eq:cosineloss}, achieved by setting all
\(\bm{e}(t)\) to the same vector for all \(t\). Because of this, the target representations are typically fixed during training.

Representation collapse can be quantified using several methods, the most prominent of which we describe below.
\paragraph{Average cosine similarity.}
Given a matrix of representations $\bm{Z}^{(L)} \in \bbR^{N \times d}$, where $\bm{Z}^{L}_{ij}$ is the activation of token $i$ at position $j$ of layer $L$, 
the average cosine similarity is: \citep{godey-etal-2024-anisotropy, tokarchuk-niculae-2024-unreasonable}. 
\begin{equation}
S_\text{avg}(\bm{Z}^{(L)}) \coloneqq \frac{2}{N(N-1)} \sum_{1 \leq i < j \leq N} \operatorname{cos}(\bm{Z}^{(L)}_i, \bm{Z}^{(L)}_j)
\end{equation}
Abnormally high cosine similarities that cannot be explained by semantic similarity may serve as a signal of complete representation collapse.

\paragraph{Matrix entropy.}
Another common approach is to examine the covariance matrix (or Gram matrix) of the representations \citep{arefin2025seqvcr, gerasimov2025fullyutilizetransformersrepresentation,razdaibiedina-etal-2023-representation,skean2024does}, defined as
$\bm{G}=\bm{Z}^{(L)}(\bm{Z}^{(L)})^\top \in \bbR^{d \times d}$.
The challenge is however to find a basis-invariant metric, since rotations of the space should lead to an identical metric value.
To this end, \citet{skean2024does} propose using the distribution of the eigenvalues of the Gram matrix. 
Denote by $\lambda_k(\bm{G})$ its $k$th eigenvalue, which measures the variance of the data along the direction spanned by the $k$th eigenvector, with $1 \leq k \leq d$.
If the data is entirely constant in some dimension $k$ then $\lambda_i(\bm{G})=0$, and implicitly the data lies exactly on a low-dimensional subspace.
More generally, since the eigenvalues are non-negative we may interpret the spectrum as a distribution by renormalizing $\tilde\lambda_k(\bm{G}) = \lambda_k(\bm{G}) / \sum_{k'} \lambda_{k'}(\bm{G})$. 
Ideally, the distribution would be close to uniform, with high entropy, so as the variance is spread evenly across all dimensions. In cases of partial collapse the entropy becomes low as the distribution peaks on a few dimensions.
This leads to the Rényi entropy of order $\alpha$ \citep{renyi1961measures} applied to the eigenvalue distribution, sometimes called the Matrix Rényi entropy \citep{Giraldo-measure-of-entropy}:
\begin{equation}
    S_{\alpha}(\bm{Z}^{(L)})=\frac{1}{1-\alpha}\log \left[ \sum_{k}\tilde\lambda_k^\alpha(\bm{G}) \right].
\end{equation}
Note that in the limit of $\alpha \to 1$, $S_\alpha$ recovers the standard Shannon entropy of the eigenvalue distribution, and indeed in this work we keep $\alpha=1$, despite the more general
definition given by \citet{skean2024does}.
In a healthy model, $S_{\alpha}$ is high, high, which indicates that the empirical covariance covers all dimensions more uniformly. In contrast, when the data only varies in a linear subspace of small dimensions (a form of dimensional collapse), $S_{\alpha}$ will be low.
\paragraph{Spherical variance.}
Another helpful way to measure representation geometry is through the distribution of the
directions of the vectors. Consider the mean vector of all (normalized) directions of a representation matrix:
\[
\bm{M}(\bm{Z}^{(L)}) = \frac{1}{N} \sum_i \frac{\bm{Z}^{(L)}_i}{\|\bm{Z}^{(L)}_i\|_2}.
\]
As this vector is the mean of points on the sphere, it will fall in the interior of the sphere (including the surface). The closer it is to the origin, the more spread out the directions are. Its length is therefore known as the spherical variance:
\citep{alma999052553502466,mardia1975statistics}
\begin{equation}
    \label{eq:svar}
    \operatorname{svar}(\bm{Z}^{(L)}) = 1-\| \bm{M}(\bm{Z}^{(L)}) \|.
\end{equation}
The main advantage of spherical variance over the methods discussed above is its computational efficiency, which makes it well suited for being reported during training.

\paragraph{Other methods.}
The token separability test, introduced by \citet{voita-etal-2019-bottom} can also be used to test if contextual representations carry distinguishing information. By taking identical tokens (\eg, the word ``is'') in various contexts, one can examine whether the model’s representations for these occurrences are different or all ``collapsed'' to the same point. This method has been adopted in recent works. For instance, \citet{gerasimov2025fullyutilizetransformersrepresentation} used a similar approach to show representation collapse in Transformer layers.

\subsection{Angular Dispersion}
\label{sec:dispersion}
Representation collapse is often measured in terms of cosine similarity for language tasks, with the underlying assumption that the directions in a $d$-dimensional space can be represented as points on the sphere $\bbS_d \subset \bbR^d$. Unlike the entirety of $\bbR^d$, the sphere is compact and has many computationally attractive properties that allow us to quantify and optimize angular dispersion. We can view optimal dispersion as a counterpart of a representation collapse on the $\bbS_d$, and finding an optimally-dispersed configuration is known as the Tammes problem \citep{tammes-1930}. It is generally not tractable in a setup with a large amount of high-dimensional points,
so several optimization approaches evolved to encourage dispersion \citep{pmlr-v119-wang20k,wang2020mmadispersion,liu2018learning,tokarchuk2025keep}.

Unlike most approaches based on pairwise distances or kernels,
\emph{Sliced dispersion} \citep{bonet2023spherical,tokarchuk2025keep} 
is an efficient alternative which avoids the quadratic complexity by making use of the fact that optimal dispersion is trivial on a circle (on $\bbS_2 \subset \bbR^2$). In this special case, any perfectly-dispersed configuration is made up of equidistant angles and given any input set of angles, the nearest dispersed configuration can be efficiently found. 
In $d=2$, given a vector of angles $\bm{\Phi}$, let the total distance between it and a perfectly-dispersed configuration of angles be written $\delta(\bm{\Phi})$.
As this distance cannot be calculated in higher dimension, we slice our data along great circles.
On a sphere $\bbS_d$, the great circles correspond to pairs of orthogonal directions \(C(\bbS_d) \coloneqq \{(\bm{P}, \bm{Q}): \bm{Q} \in \bbS_d, \bm{Q} \in \bbS_d, \DP{\bm{P}}{\bm{Q}}=0\}\).
Let $\bar{\bm{Z}}$ denote a configuration of directions (\eg, normalized representation vectors).
If we denote by $\bar{\bm{Z}}_{\bm{PQ}}$ 
the projection of the directions in $\bar{\bm{Z}}$ 
onto the great circle $(\bm{P},\bm{Q})$,
sliced dispersion optimizes
\begin{equation}
R_\text{sliced}(\bar{\bm{Z}}) \coloneqq \bbE_{\bm{P,Q}} \left[ \delta(\bar{\bm{Z}}_{\bm{PQ}})\right],
\end{equation}
where the expectation is over the uniform distribution on the $C(\bbS_d)$.
In words, this objective minimizes the expected distance to an optimally-sliced configuration along any great circle.

\subsection{Efficiency}
\label{sec:efficiency_background}
While recent LLMs have changed the machine learning landscape, their performance comes with a trade-off in size and energy usage \citep{shterionov2023ecological}, as well as the need for a wireless connection to run models in the cloud. Smaller models are still preferable where they need to be run on edge devices, have a lower energy footprint, may need to be run on devices without a wireless connection, or need to be trained on a smaller set of private data \citep{menghani2023efficient}. There are several ways to increase the efficiency of a model, without necessarily sacrificing its performance, either during or after training, during inference, or both. These include such methods as distillation, pruning, and, the focus here, quantization \citep{menghani2023efficient}. 
One particular way of doing quantization involves compressing the model by modifying the data type from (typically) \texttt{float32} precision to a lower precision, such as half-precision floating-point \texttt{float16} or \texttt{int8} \citep{tang2024survey}. Since quantization necessarily involves representing less information, it must strike a balance between efficiency and performance \citep{menghani2023efficient}. Despite its usefulness in increasing training efficiency and widespread application for training large neural networks \citep{grattafiori2024llama3herdmodels,alves2024tower}, \citet{barbero2024transformers} show that low-precision training further amplifies the issue of representation collapse. Quantization in this space could push the model to use the low-precision space more effectively.

\section{Regularized Transformer}
In order to prevent representation collapse and promote diversity in NMT representations, we propose simple yet efficient regularization on top of the next token prediction objective shown in \Cref{eq:discrete} \begin{equation}\label{eq:rnmt}
\begin{gathered}
   L_\text{RNMT}(y_i=t; \bm{y}_{<i}, \bm{x}) = L_\text{NMT} +\gamma R(\bm{Z}^{L}),
\end{gathered}
\end{equation}
where $R(\bm{Z}^{L})$ is a sliced dispersion regularization  discussed in \Cref{sec:dispersion} over an output of a layer $Z^{L}$ and $\gamma$ is a dispersion weight. $R(\bm{Z}^{L})$ can be, in principle, any latent representation of the source and target sequences. We propose to focus on the decoder's latent representations in the transformer, namely the decoder output and decoder embeddings.  \citet{gerasimov2025fullyutilizetransformersrepresentation} showed recently that deeper layers of Transformers exhibit a higher degree of collapse. Therefore, we will apply regularization to the decoder output $\bm{H}$. \citet{tokarchuk-niculae-2024-unreasonable} showed that the embeddings of rare tokens collapse to similar representations, making it difficult to distinguish between them, so we apply dispersion on top of the embeddings representation $\bm{E}$. Additionally, we propose to apply dispersion on the encoder output representation $\bm{F}$.

Similar to the CoNMT case, regularization is applied to the target embeddings, as our primary goal is to prevent the collapse of target representations.

\begin{equation}\label{eq:rconmt}
\begin{gathered}
   L_\text{RCoNMT}(y_i=t; \bm{y}_{<i}, \bm{x}) = L_\text{CoNMT} \\+\gamma R(\bm{Z}^{(l)}=\bm{E}).
\end{gathered}
\end{equation}

Note that calculating regularization over the embedding matrix $\bm{E}$ can be costly. Therefore, given the insights from previous studies \citep{tokarchuk-niculae-2024-unreasonable}, we randomly subsample from the pool of rare tokens, \ie, embeddings with a rank higher than the half of the vocabulary size.

\section{Experimental Results}
\begin{figure*}[t]
    \centering
     \begin{subfigure}[b]{\linewidth}
    \includegraphics[width=\linewidth]{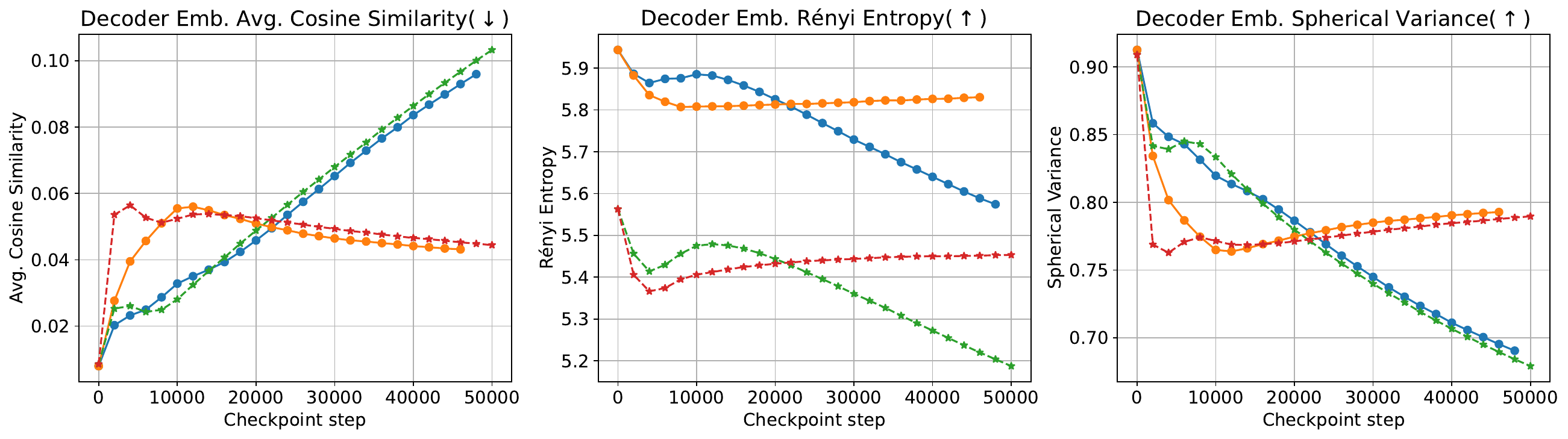}
    \end{subfigure}
    \hfill
    \begin{subfigure}[b]{\linewidth}
            \includegraphics[width=\linewidth]{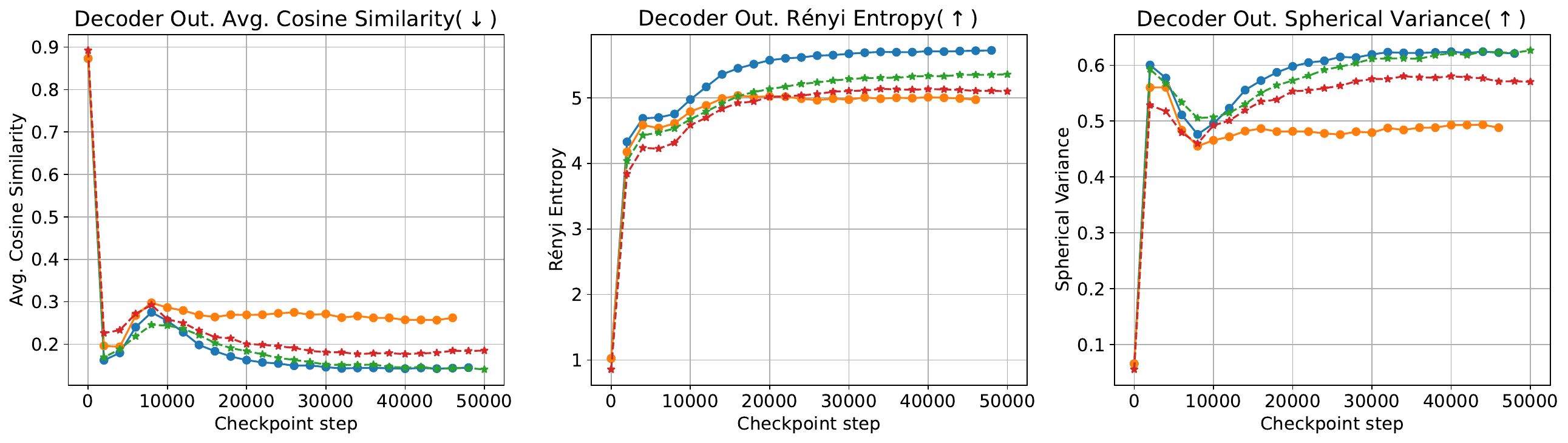}
    \end{subfigure}
    \hfill
    \begin{subfigure}[b]{\linewidth}
            \includegraphics[width=\linewidth]{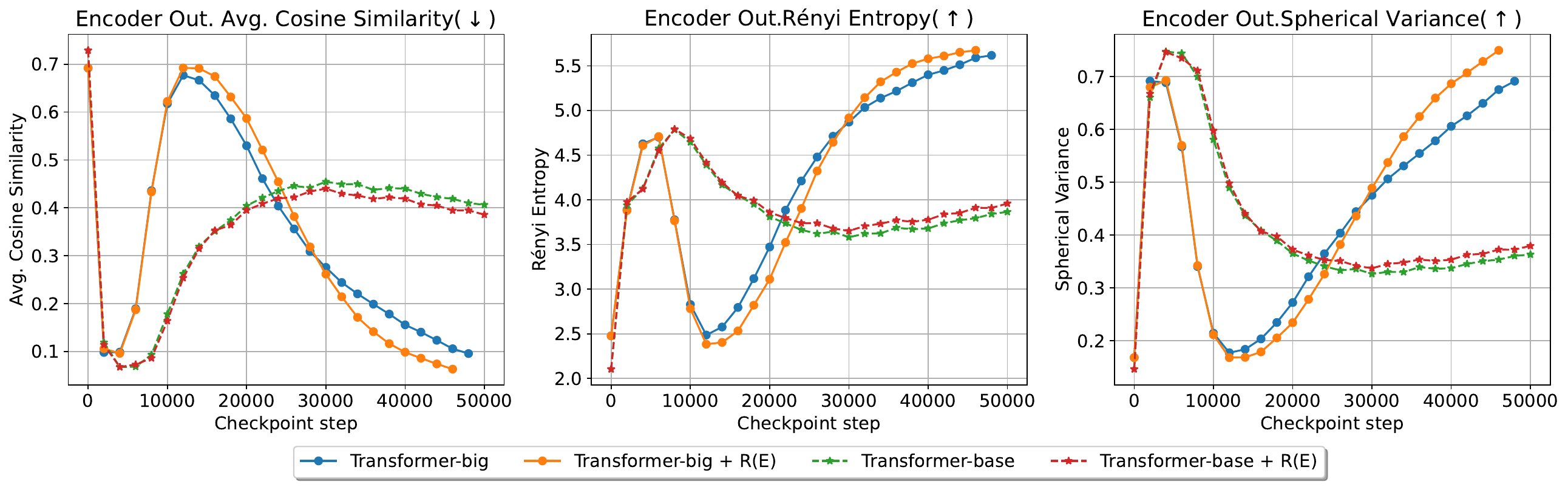}
    \end{subfigure}
    \caption{\label{fig:collapse-nmt}Average cosine similarity, Rényi entropy and spherical variance for decoder output, decoder embeddings and encoder outputs for \tbig{} and \tbase{} models.}
\end{figure*}
\subsection{Data and Evaluation}
 We provide results on the English$\leftrightarrow$German language pair in both directions, namely \langpair{en}{de} and \langpair{de}{en}. We use 34M cleaned training sentences from publicly available WMT19 dataset \citep{barrault-etal-2019-findings} provided by HuggingFace\footnote{\url{https://huggingface.co/datasets/wmt/wmt19}}, and tokenize it with BPE subword units\footnote{\url{https://github.com/glample/fastBPE}} \citep{sennrich-etal-2016-neural}. We evaluate the models using \texttt{sacrebleu} \citep{post-2018-call} and COMET \citep{rei-etal-2020-comet} on \texttt{newstest2018} and \texttt{newstest2019}. Statistics of the training and test data can be found in \Cref{tab:data-stats} of the \Cref{app:data-stats}. We additionally report results for the Romanian$\rightarrow$English language pair in \Cref{app:additional-results-roen}.

\subsection{Models and Training}
We study representation collapse in the \texttt{Transformer-big} and \texttt{Transformer-base} models defined by \citet{Vaswani-trafo}, using the \texttt{fairseq} framework \citep{ott2019fairseqXXXX}. Additionally, we train a CoNMT model with the same architecture as \tbig{}. Each model is trained for 50,000 steps on a single NVIDIA H100 GPU using the Adam optimizer \citep{Kingma2014AdamAM}. The learning rate is set to $5\cdot10^{-4}$ with 10,000 warm-up steps. The weight of the $\gamma$ parameter is tuned on the development set for each component based on the \tbase{} model in the range $\{10^k : k \in \{-2, -1, 0, 1, 2\}\}$. We used Sliced regularizer for all models with regularization due to its efficiency.

\subsection{Post-training quantization}
\label{subs:quant_method}
As applying dispersion to a model with training-aware quantization is somewhat involved, we instead opt for post-training quantization on the \tbig{} and the \tbig{} with dispersion. We apply post-training quantization using the \texttt{CTranslate2} framework, introduced by \citet{klein2020efficient}. It allows for quantization to various compression levels, and applies other efficiency techniques, such as weights pre-packing, to allow for a fast runtime after quantizing the model. Quantization may either be performed for one compression type, e.g. \texttt{float16}, in which case both the weights are stored and all layers are run at that precision level. In a mixed precision setting, e.g. \texttt{int8\_float16}, the weights of the embeddings and linear layers are quantized to \texttt{int8}, while the rest of the layers are run in \texttt{float16}. We quantized the \tbig{} model to \texttt{float16} and \texttt{int8\_float16}.\footnote{\url{https://opennmt.net/CTranslate2/quantization.html}} The \texttt{float16} model acts as a comparison to the quantization-aware trained \tbig{} model. Because the \tbase{} model is approximately four times smaller  in terms of parameters than the \tbig{} model, we included \texttt{int8\_float16} as a way to compare two models with a similar size overall. See \Cref{app:quant-deen} for a size comparison of the quantized models on disk. We run inference on the GPU.

\subsection{Representation Collapse in NMT}

First, to identify the collapse phenomenon, we examine how the collapse metrics defined in \Cref{sec:nmt-collapse} evolve over the course of training for the \langpair{de}{en} language pair. For each batch of 4,096 tokens, we compute the matrix-based Rényi entropy, spherical variance, and average cosine similarity for the decoder output $\bm{H}$, decoder embeddings $\bm{E}$, and encoder output $\bm{F}$. \Cref{fig:collapse-nmt} presents the averaged metric values across all batches at each training step for the \tbig{}, \tbig{}+dispersion (embeddings), and \tbase{} models.

The \tbase{} model exhibits stronger collapse in the encoder output and decoder embeddings compared to the \tbig{} model. Interestingly, the decoder output does not display clear signs of collapse in either configuration. While the \tbig{} encoder output partially recovers during training, the \tbase{} encoder representations collapse more severely and remain less diverse. We hypothesize that dimensionality and dispersion have strong connection with each other, and larger dimensionality naturally promotes dispersion, which is also in line with observations in previous works \citep{tokarchuk2025angular}.

Applying dispersion regularization to the \tbig{} embeddings improves all collapse metrics. However, applying dispersion to the decoder embeddings adversely affects the representation diversity of the decoder output, as reflected by higher (average) cosine similarity and lower entropy and variance scores. 

\begin{table*}[]
    \centering
    \small
    \setlength{\tabcolsep}{3pt}
    \begin{tabular}{lcccccccccc}
    \toprule
         \multirow{3}{*}{model} & \multicolumn{4}{c}{\texttt{\langpair{en}{de}}} && \multicolumn{4}{c}{\texttt{\langpair{de}{en}}}\\
          & \multicolumn{2}{c}{\texttt{newstest18}} & \multicolumn{2}{c}{\texttt{newstest19}} && \multicolumn{2}{c}{\texttt{newstest18}} & \multicolumn{2}{c}{\texttt{newstest19}}\\ 
          & BLEU & COMET & BLEU & COMET && BLEU & COMET & BLEU & COMET\\
          \midrule
          \tbig{} &41.9&0.803&31.7&0.790&&43.0&0.811&39.3&0.776\\
          
          \quad + $R(\bm{H})$ &41.6&0.803&31.7&0.789 && 42.5 &0.808&39.0&0.774\\
          \quad + $R(\bm{E})$ & \textbf{42.6} &\textbf{0.806}&\textbf{32.6}& \textbf{0.793} && 43.2 & 0.812 & 39.2 & 0.776\\
          \midrule
          \tbase{} & 40.9 & 0.792& 30.9 & 0.782 & & 41.7 & 0.799& 37.8 & 0.762\\
          \quad + $R(\bm{H})$ & 40.4 &0.788&30.6&0.777 && 41.0 & 0.797&37.5 & 0.759\\
          \quad + $R(\bm{E})$ &41.2 &0.795&31.2&0.783 && 41.5 & 0.801 & 37.8 & 0.762\\ %
           \quad + $R(\bm{F})$ &40.7&0.791&31.1&0.781&&42.0&0.801&37.8&0.764\\
    \bottomrule
    \end{tabular}
    \caption{\label{tab:main_res}BLEU and COMET on \texttt{newstest18} and \texttt{newstest19} for \langpair{en}{de} end \langpair{en}{de} with different dispersion regularizers. Values in bold indicate results that are statistically significant compared to the baseline (p-value < 0.05) }
\end{table*}

\paragraph{Translation Quality.}
\begin{figure}[ht]
    \centering
    \includegraphics[width=.8\linewidth]{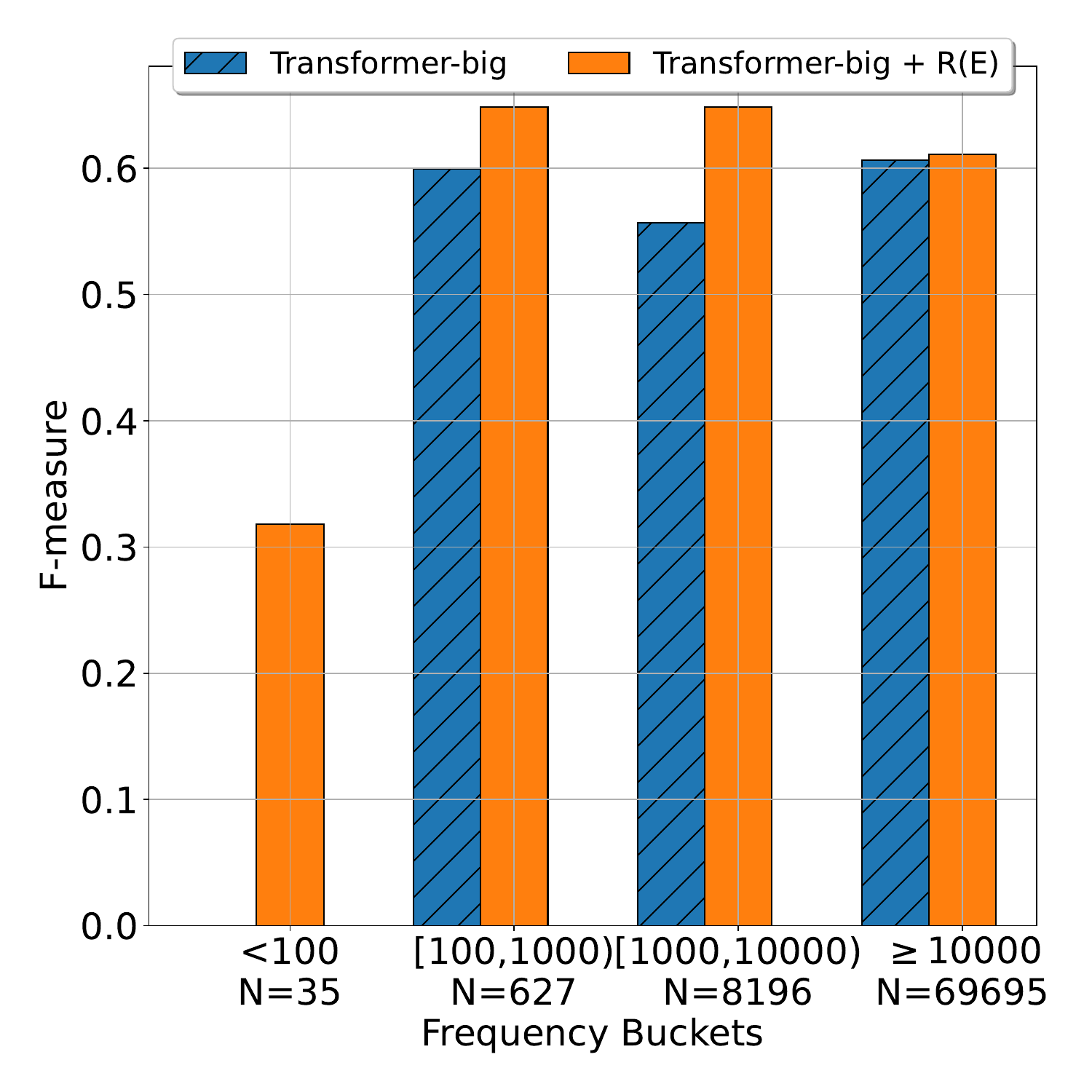}
    \caption{\label{fig:fmes-ende} F1 score for each vocabulary token's frequency bucket for \langpair{en}{de}. Note that the F1 score for tokens with frequency $< 100$ for the model without regularization is 0. }
\end{figure}
In \Cref{tab:main_res}, we quantify the effect of dispersion regularization applied to different components of the model using standard translation quality metrics. Results for both translation directions show that applying dispersion to the decoder embeddings can further improve the performance of the strong \tbig{} baseline, consistent with the trends observed in \Cref{fig:collapse-nmt}. Furthermore, applying dispersion regularization directly to the encoder embeddings of the \tbase{} model effectively mitigates encoder-side representation collapse.

We analyze translation quality at the token level using \texttt{compare-mt} \citep{compare-mt}. \Cref{fig:fmes-ende} presents the F1 score across token frequency buckets on the development set of the \langpair{en}{de} model. The model with dispersion applied to the decoder embeddings clearly outperforms the non-dispersed baseline on rare tokens (frequency below 100), for which the latter achieves an F1 score close to zero. This improvement on rare tokens also translates into better overall performance. A similar trend is observed for the \langpair{de}{en} direction, although the effect is less pronounced due to the smaller number of rare tokens. For detailed analysis on \langpair{de}{en}, please refer to \Cref{app:freq-analysis-deen}.

\begin{figure}
    \centering
    \includegraphics[width=\linewidth]{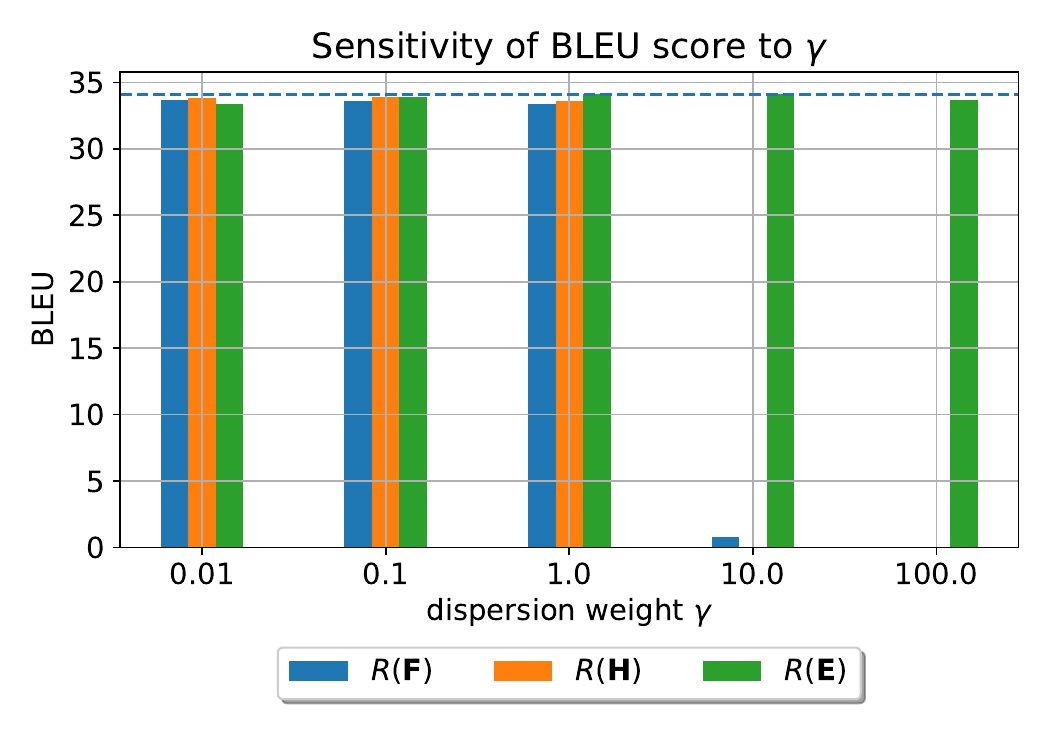}
    \caption{\label{fig:fig:dispersion-weight} BLEU score for the \langpair{de}{en} development set \texttt{newstest17} of \tbase{} model for different parameters $\gamma$. Horizontal blue line indicates the BLEU score of the unregularized \tbase{} model.}
\end{figure}

\paragraph{Sensitivity to dispersion weight}
Varying the parameter $\gamma$ in \Cref{eq:rnmt} strongly influences the degree of dispersion and, consequently, the model’s performance. We conduct a sensitivity analysis on the \tbase{} model for the \langpair{de}{en} language pair across different model representations, with results shown in \Cref{fig:fig:dispersion-weight}. Our experiments indicate that the decoder embeddings are less sensitive to the dispersion weight, whereas the encoder output is more susceptible to over-dispersion, which can degrade performance.

\begin{figure*}
    \includegraphics[width=\linewidth]{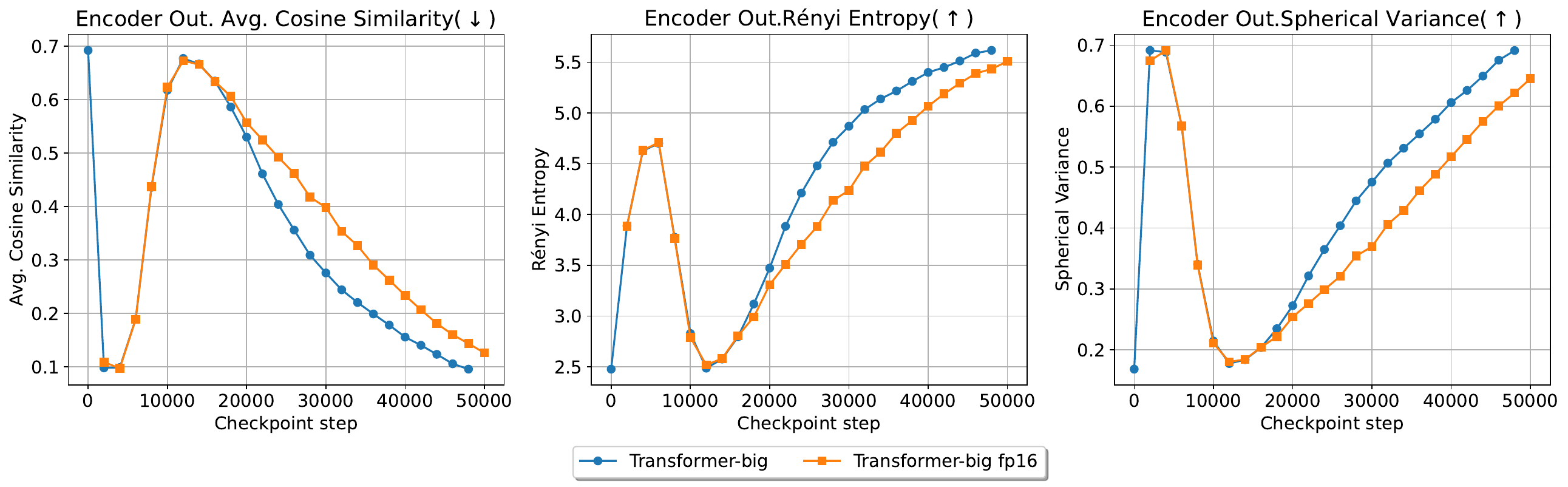}
    \caption{\label{fig:fp16enc} Comparison of the encoder outputs representation collapse metrics for \tbig{} and \tbig{} \texttt{fp16} \langpair{de}{en}.}
\end{figure*}

\paragraph{Quantization.}
To examine the model’s behavior with respect to representation collapse under half-precision training, we train the \tbig{} model in \texttt{fp16}. There is no noticeable differences in the representations of the decoder output or decoder embeddings between \tbig{} and \tbig\texttt{-fp16} models. Encoder output metrics, however, indicate a slightly higher degree of collapse for \texttt{fp16} (\Cref{fig:fp16enc}).

From \Cref{tab:post_train_q} we can see that the both post-training quantization of the \tbig{} model in \texttt{float16} and \tbig{} model trained in \texttt{float16} perform similarly to the full precision \tbig{} model. However, quantization with \texttt{int8\_float16} hurts both models performance significantly, without clear evidence that dispersion of the decoder embeddings helps to preserve representation better. Even so, the benefit of dispersion is preserved even after quantization. Nonetheless, investigating representation collapse under reduced precision warrants further and more systematic study, especially in a dynamic setting.

\begin{table}[ht]
    \centering
    \small
    \setlength{\tabcolsep}{3pt}
    \begin{tabular}{llcc}
    \toprule
      prec. &reg. &BLEU & COMET  \\ \midrule
      \texttt{float32} & \xmark & 31.7 & 0.790 \\
       \texttt{float32} & $R(\bm{E})$ & 32.6 & 0.793 \\
       \texttt{float16}* & \xmark & 31.7 & 0.787 \\
      \midrule
      \texttt{float16} & \xmark & 31.7 & 0.845 \\
       \texttt{float16} & $R(\bm{E})$ & 32.6 & 0.847\\
       \texttt{int8\_float16} & \xmark &  27.5  & 0.745 \\
    \texttt{int8\_float16} & $R(\bm{E})$ &  28.0 & 0.813 \\
      \bottomrule
    \end{tabular}
    \caption{\label{tab:post_train_q}Comparison of quantization-aware training (\texttt{float*}) and post-training quantization on the \tbig{} \langpair{en}{de} model with and without dispersion regularization on \texttt{newstest19}.}
\end{table}

\subsection{Full Collapse in CoNMT}
\begin{figure*}[ht]
    \includegraphics[width=\linewidth]{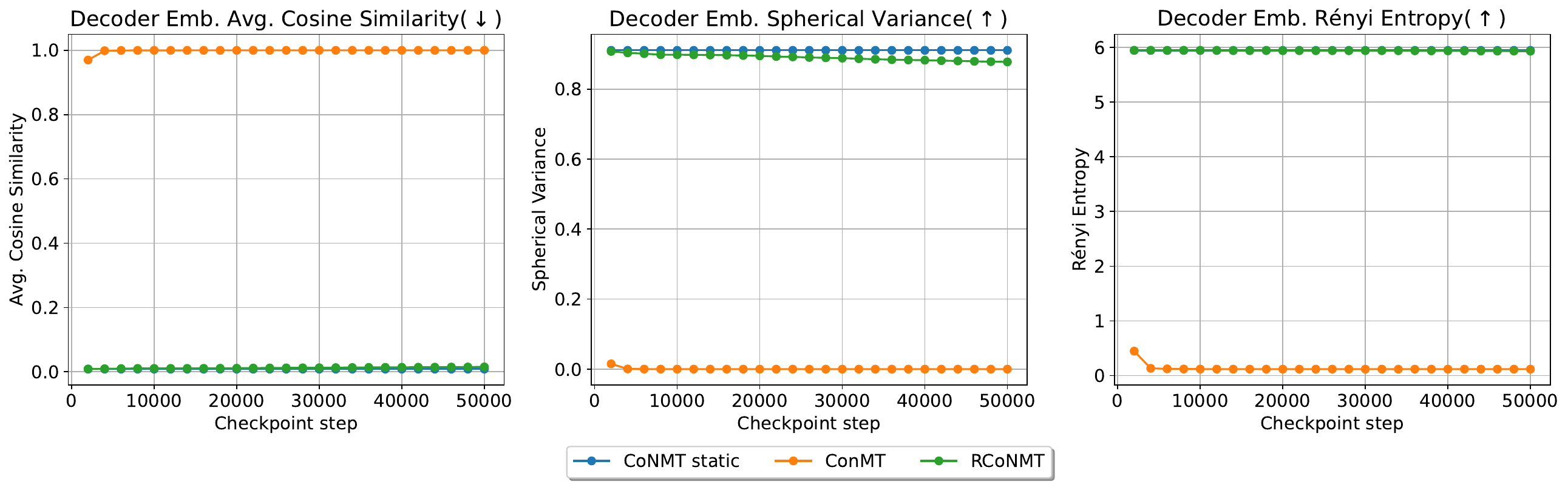}
    \caption{\label{fig:conmt-collapse} Representation collapse in target representations of the CoNMT models. CoNMT static refers to the model trained with frozen target embeddings.}
\end{figure*}

\Cref{tab:conmt} demonstrates that dispersion regularization effectively mitigates representation collapse during end-to-end training of CoNMT. Notably, without applying regularization to the embeddings, the model collapses to a trivial solution and fails to achieve a BLEU score above zero unless the embeddings are frozen.
\begin{table}[ht]
    \centering
    \small
    \setlength{\tabcolsep}{3pt}
    \begin{tabular}{llccc}
    \toprule
         objective & init. & train $\bm{E}$ & BLEU & COMET \\
          \midrule
          $L_{\text{CoNMT}}$ & rand. & \xmark &33.9&0.713\\
        $L_{\text{CoNMT}}$  & rand. & \cmark & 0.0 & 0.000 \\ \midrule
           $L_{\text{CoNMT}}$  & NMT & \xmark &32.9&0.704\\
           $L_{\text{CoNMT}}$  & NMT & \cmark & 0.0 & 0.000\\
           $L_{\text{CoNMT}}$  & RNMT & \xmark & \textbf{36.6} & \textbf{0.749}\\
           $L_{\text{CoNMT}}$  & RNMT & \cmark &0.0&0.000 \\ \midrule
           $L_{\text{RCoNMT}}$  & rand. & \cmark &33.2&0.708\\ 
          $L_{\text{RCoNMT}}$& NMT & \cmark &31.3&0.700\\
          $L_{\text{RCoNMT}}$ & RNMT & \cmark &30.9&0.694\\
    \bottomrule
    \end{tabular}
    \caption{\label{tab:conmt}\langpair{de}{en} translation scores on \texttt{newstest2019}.
    }
\end{table}

Looking more closely at the dynamics of representations, we can clearly see a complete representation collapse in \Cref{fig:conmt-collapse}, where the spherical variance and matrix-based entropy drop to 0, and the average cosine similarity reaches 1. The model with regularization successfully prevents this collapse, achieving stability comparable to the model with fixed (frozen) embeddings. Note, that changing initialization of target representation from unregularized model or random embeddings to the embeddings from regularized model achieves the best overall performace for CoNMT.

\section{Conclusion}
In this work, we investigated the phenomenon of representation collapse in neural machine translation. Through experiments with the Transformer-big model, we demonstrated that promoting dispersion in representations can further enhance an already strong baseline. We observe that this benefit holds even after quantizing the model. Moreover, our results indicate that dispersion plays a critical role in preventing full collapse in continuous-output NMT, highlighting its potential as a simple yet effective strategy for stabilizing representation learning in Transformer-based models.

\section*{Limitations}
\paragraph{Limited evaluation scenarios.}In this work, we focused on the representations of three specific layers within the Transformer model and proposed a regularization strategy to prevent collapse. However, representation collapse may also occur in other layers, and further investigation would provide a more comprehensive understanding of the phenomenon. Moreover, we applied only one regularization component at a time. Our experiments suggest that regularizing one layer can negatively affect the representational diversity of another. A combined or coordinated regularization approach across multiple layers may therefore offer a more effective solution and warrants further study.
\paragraph{Low-resource tasks.}While WMT19 English–German is a widely used translation benchmark, our results indicate that dispersion yields the greatest benefits for rare tokens, an effect that is likely to be more pronounced in low-resource language pairs. Extending our approach to such settings, as well as to multilingual translation scenarios, would provide stronger empirical support for the proposed method.

\section*{Acknowledgements}
We thank members of the UvA Language Technology Lab for discussion and feedback on the manuscript. This work is supported by the Dutch Research Council (NWO) via VI.Veni.212.228. Maya K. Nachesa is supported by the ROBUST project (number KICH3.LTP.20.006) which is (partly) financed by the Dutch Research Council (NWO), RTL, and the Dutch Ministry of Economic Affairs and Climate Policy (EZK) under the program LTP KIC 2020-2023. Sergey Troshin is partially supported by ``Hybrid Intelligence: augmenting human intellect'' (\url{https://hybrid-intelligence-centre.nl}) with project number 024.004.022 of the research program ``Gravitation'' which is (partly) financed by the Dutch Research Council (NWO). The authors also thank SURF (\url{www.surf.nl}) for the support in using the National Supercomputer Snellius. 

\bibliography{custom}
\appendix
\section{Training Details}
\subsection{Data Statistics}\label{app:data-stats}

\begin{table}[ht]
\centering
\resizebox{0.8\linewidth}{!}{%
\begin{tabular}{lcccc}
\hline
\toprule
&lang.& \#sent. & \#toks. & \#vocab.\\  \midrule 
train & \texttt{de}& 34M & 916M &\multirow{4}{*}{42024}\\ 
{\texttt{newstest2017}} &\texttt{de} & 3004 & 78553 &\\
{\texttt{newstest2018}}& \texttt{de} & 2998 & 81875 & \\
\texttt{newstest2019}& \texttt{de} & 2000 & 47035 &  \\\midrule
train & \texttt{en}& 34M & 888M & \multirow{4}{*}{42024}\\ 
{\texttt{newstest2017}}& \texttt{en} & 3004  & 75011 & \\
{\texttt{newstest2018}} &\texttt{en} & 2998& 78925& \\
\texttt{newstest2019}& \texttt{en} & 2000 & 46261 & 
\\\bottomrule
\end{tabular}%
}
\caption{\label{tab:data-stats} Data statistics for training and test splits.}
\end{table}

\section{Additional Results}
\subsection{Frequency-based Analysis}\label{app:freq-analysis-deen}
\begin{figure}
    \centering
    \includegraphics[width=0.7\linewidth]{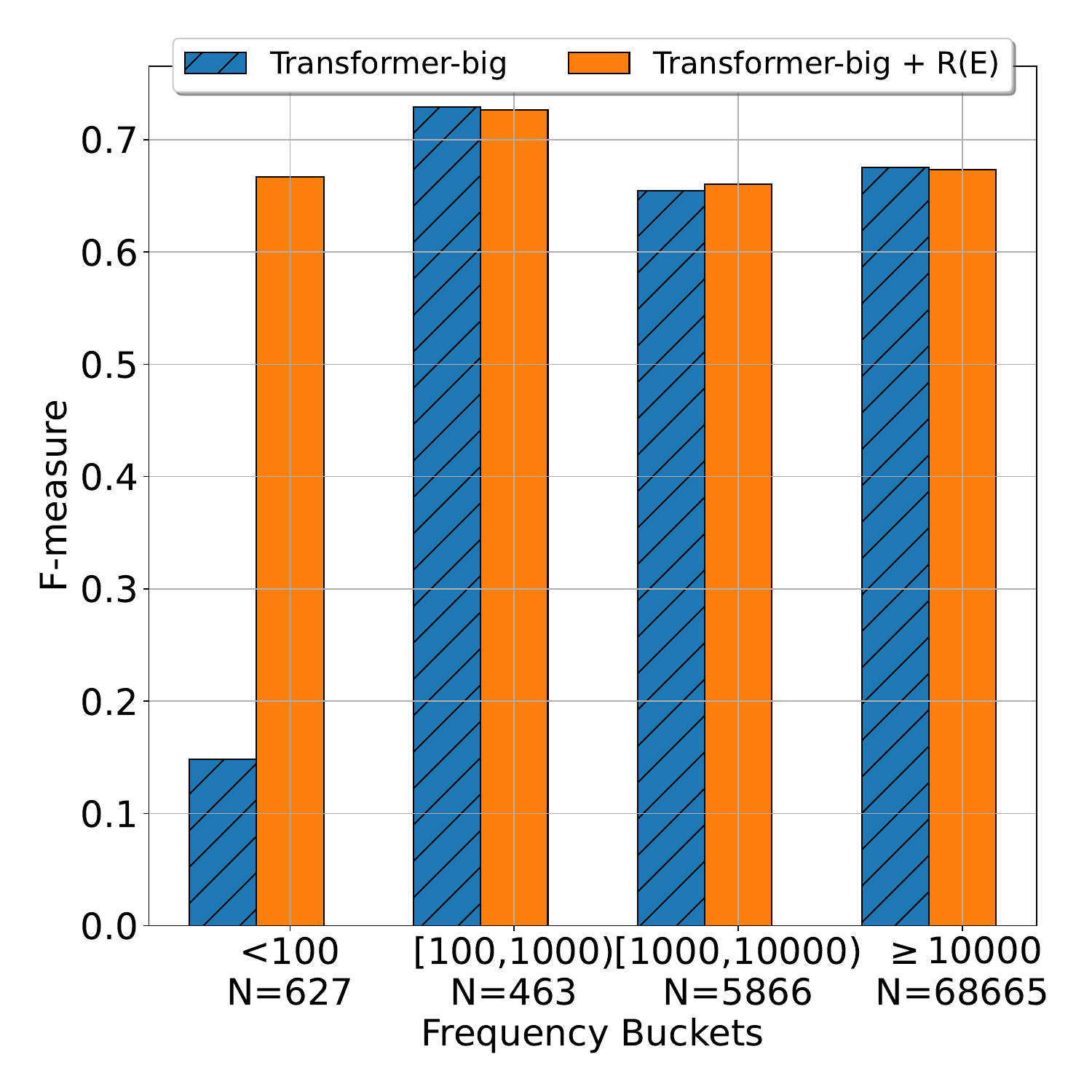}
    \caption{\label{fig:fmeas-deen}F1 score for \langpair{de}{en} for each vocabulary token's frequency bucket.}
\end{figure}

\Cref{fig:fmeas-deen} provides frequency-based token analysis for \langpair{de}{en} \tbig{} model. Similar to the \Cref{fig:fmes-ende}, the performance of the model with dispersion regularization on rare tokens is significantly better than the baseline model.

\subsection{Quantization}\label{app:quant-deen}
\begin{table}[ht]
    \centering
    \small
    \setlength{\tabcolsep}{3pt}
    \begin{tabular}{llr}
    \toprule
    architecture & precision & size \\
    \midrule
    transformer-big & \texttt{float32} & $100\%$ \\
    transformer-big & \texttt{float16} & $50\%$ \\
    transformer-big & \texttt{int8\_float16} & $25\%$\\
    transformer-base & \texttt{float32} & $30\%$ \\
    \bottomrule
    \end{tabular}
    \caption{\tbig{} size on disk for the original models and quantized versions. Percentages based on disk storage.}
\label{tab:quantization_model_size}
\end{table}

\Cref{tab:quantization_model_size} shows the model size on disk both for the original \tbig{} model, as well as various compression levels in CTranslate2. This was done as a proxy for parameter counts, as CTranslate2 returns a compressed model file. The \texttt{float32} model compressed with CTranslate2 is included as a reference point, such that the differences in file size are only due to the compression level. Note that the \texttt{int8\_float16} model is approximately four times smaller than the \texttt{float32} compressed model. This is roughly equivalent to the difference in size between a \tbase{} and a \tbig{}.

\section{Additional Results}
\label{app:additional-results-roen}
We report additional results for model training with dispersion regularization on the WMT16 Romanian→English (\langpair{ro}{en}) benchmark. The dataset contains 612K training sentence pairs. For subword tokenization, we use the same SentencePiece model \citep{kudo-richardson-2018-sentencepiece} across all language pairs, specifically the one employed in the mBART multilingual model \citep{liu-etal-2020-multilingual-denoising}. The vocabulary size is 27K on the target side and 28K on the source side. All hyperparameters are identical to those used for \langpair{de}{en}, except for dropout, which is set to 0.3. 
\begin{table}[ht]
    \centering
    \small
    \setlength{\tabcolsep}{3pt}
    \begin{tabular}{lccccccc}
    \toprule
         \multirow{3}{*}{model} & \multicolumn{4}{c}{\texttt{\langpair{ro}{en}}}\\
          & \multicolumn{2}{c}{\texttt{newsdev16}} & \multicolumn{2}{c}{\texttt{newstest16}}\\ 
          & BLEU & COMET & BLEU & COMET \\
          \midrule
          \tbig{} &32.6&0.786&31.2&0.790\\
          
          \quad + $R(\bm{H})$ &32.7&0.785&31.0&0.788 \\
          \quad + $R(\bm{E})$ & 33.0 &0.787&31.4&0.791  \\
          \midrule
          \tbase{} & 33.2 & 0.793& 31.8 & 0.795  \\
          \quad + $R(\bm{H})$ & 33.0 &0.789&31.9&0.791 \\
          \quad + $R(\bm{E})$ & \textbf{33.6} & 0.794&\textbf{32.3}&\textbf{0.799} \\ %
           \quad + $R(\bm{F})$ &33.1&0.789&32.1&0.793\\
    \bottomrule
    \end{tabular}
    \caption{\label{tab:roen_res}BLEU and COMET on \texttt{newsdev} and \texttt{newstest} for \langpair{ro}{en} with different dispersion regularizers. Values in bold indicate results that are statistically significant compared to the baseline (p-value < 0.05) }
\end{table}
\Cref{tab:roen_res} shows that the \tbase{} model benefits substantially from embedding dispersion, whereas for \tbig{} only marginal improvements are observed. Overall, these observations are consistent with our main results and further support the conclusion that dispersion is a beneficial property for the model.
\end{document}